# Parallel Dither and Dropout for Regularising Deep Neural Networks

Andrew J.R. Simpson [#1]

[#] *Centre for Vision, Speech and Signal Processing, University of Surrey*
*Surrey, UK*
[1] `Andrew.Simpson@Surrey.ac.uk`

*Abstract*—**Effective regularisation during training can mean the difference between success and failure for deep neural networks. Recently, *dither* has been suggested as alternative to *dropout* for regularisation during batch-averaged stochastic gradient descent (SGD). In this article, we show that these methods fail without batch averaging and we introduce a new, parallel regularisation method that may be used without batch averaging. Our results for parallel-regularised non-batch-SGD are substantially better than what is possible with batch-SGD. Furthermore, our results demonstrate that dither and dropout are complimentary.**

## I. INTRODUCTION

Whilst applied research in deep neural networks (DNN) has progressed rapidly in recent times, not least due to progressive increases in computing capacity, relatively little progress has been made on the fundamentals of how DNN learn. In particular, the role and mechanisms of regularisation have received little attention.

The most popular current regularisation method, known as *dropout* [1], is practically ubiquitous and tends to accompany batch-averaged stochastic gradient descent (SGD). *Dither* [2] has also been demonstrated to be equal or better as regulariser in the same batch-averaged context. However, little or no work has gone into regularisation for non-batch-based methods. To some extent this is probably because non-batch-averaged SGD tends to work much less well than batch-averaged SGD and typically the available regularisation methods do not help much or at all because they appear to rely upon batch averaging (or other integration methods such as momentum) to produce reliable estimated gradients.

Generally, then, there are several open questions for regularisation. Given that dropout and dither have both been demonstrated to be similarly effective regularisers, but with different mechanisms (dropout is multiplicative and dither is additive) and somewhat different performance (dither appears to produce more rapid learning [see 2]), it may be that the two mechanisms address different components of the regularisation problem. In particular, dither addresses the nonlinear distortion components produced by the activation functions in a DNN [2]. On the other hand, dropout appears to address architectural dependencies between neurons [1] and so it may be that a combination of the two will be useful. Or, to rephrase, dither addresses nonlinear distortion in the activations whilst dropout addresses potentially independent issues in the weights. Furthermore, it has also been suggested that batch-averaging, mediated by the data itself, acts as a regularisation in its own right [2]. Hence, it is important to evaluate the regularisation effects of dither and dropout independently of batch averaging and it is also important to develop a regularisation method capable of regularising non-batch SGD.

In this paper, we describe a parallelised regularisation method designed for non-batch SGD. Our results demonstrate that non-batch SGD with parallel dither and dropout regularisation out-performs the best batch-averaged equivalent.

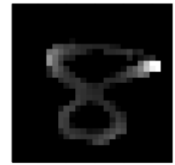

Fig. 1. **Example MNIST image.** We took the 28x28 pixel images and unpacked them into a vector of length 784 to form the input at the first layer of the DNN.

## II. METHOD

For case study, we use the well-known computer vision problem of hand-written digit classification using the MNIST dataset [3]. For the input layer we unpacked the images of 28x28 pixels into vectors of length 784. An example digit is given in Fig. 1. Pixel intensities were normalized to zero mean. Replicating Hinton's [4] architecture, but using the biased sigmoid activation function [2], we built a fully connected network of size 784x100x10 units, where the 10-unit softmax output layer corresponds to the 10-way digit classification problem.

Operating within the so-called 'small-data regime' (as in [2]), we used only the first 256 training examples of the MNIST dataset and tested on the full 10,000 test examples.

We trained several instances of the model with non-batch SGD (equivalent to a batch size of 1 in batch-averaged SGD). The first was a baseline model without regularisation. The second was the baseline model regularised with dropout. The third was the baseline model regularised with dither. The fourth was the baseline model regularised in parallel using dropout. The fifth was the baseline model regularised using parallel dither and the final instance was the baseline model regularised using a combination of parallel dither and parallel dropout.

*Parallel dither and dropout*. During non-batch SGD, each training example was replicated 100 times to form a parallel set. For parallel dither, each element of this set was dithered independently by adding uniform random noise of zero mean and unit scale and the gradients computed for each element independently. For parallel dropout, each element of the parallel set was subject to different (random) dropout during gradient computation to provide an equivalent set of gradients. For parallel dither w/ dropout, both dither and dropout were applied at the same time (i.e., the parallel set was still of size: 100). Then, each parallel set of gradients (representing a single training example) was averaged and applied. Batch averaging across training examples was not applied.

Each separate instance of the model was trained for 100 full-sweep iterations of non-batch SGD (without momentum) and the test error computed (over the 10,000 test examples) at each iteration. For reliable comparison, each instance of the model was trained from the exact same random starting weights. A learning rate (SGD step size) of 1 was used for all training. All dropout was at the 50% level. These parameter choices allow direct comparison with the batch-SGD results of [2].

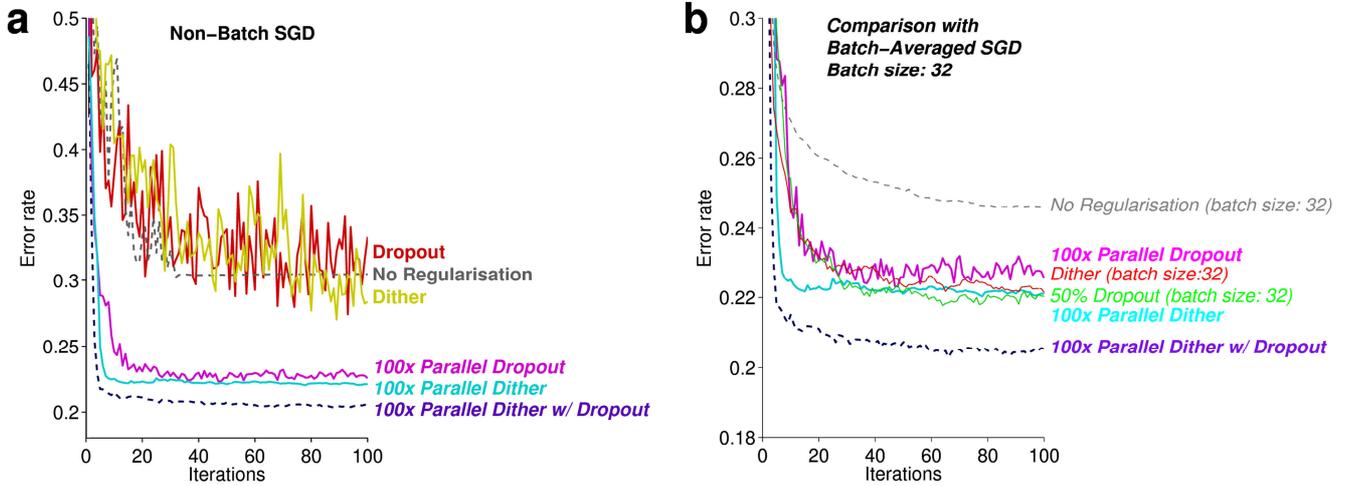

**Fig. 2. Parallel regularisation with dither and dropout. a** plots the test error function of SGD iterations, for the various non-batch SGD models **b** plots the same parallel model data of panel a compared with the best batch-averaged results of [2] for comparison. *Note: 1) the y-axis in both panels is somewhat cropped for better scale after the first iteration of training and 2) the y-axis in panel **b** is zoomed in for easier comparison and hence is not plotted at the same scale as the same 100x parallel regularised data as plotted in panel **a**.*

## III. RESULTS

Fig. 2a plots the test-error rates, as a function of full-sweep SGD iterations, for the various non-batch SGD trained models. The non-batch model trained without regularisation performs poorly and things don't improve with either dither or dropout. This is not surprising since both dropout and dither require integration over batches for their estimated gradients to be useful.

The parallel regularised models are also plotted in Fig. 2a. The parallel version of dropout performs substantially better and demonstrates useful regularisation. The parallel version of dither performs better than the parallel dropout and the faster learning rate of dither is consistent with the batch-SGD results of [2]. Finally, the combination of parallel dither with parallel dropout results in the best performance; learning rate is very rapid indeed and ultimate performance is improved by a good margin. The fact that the parallel combination performs better than either individual parallel method tends to suggest that, as anticipated, the two methods do indeed address different aspects of DNN regularisation. Presumably then, the combined advantage results from dither acting on the quality of the activations and dropout acting on the quality of the weights.

Fig. 2b plots a comparison of the parallel regularised models with the best batch-SGD results from [2] where a batch size of 32 was found to be optimal using conventional dropout or dither. *Note that the data, model architecture, hyper-parameters and random starting points are identical across the experiments of this article and* [2]. Compared to the dither or dropout regularised batch-SGD (batch size: 32) results from [2], parallel dropout (non-batch-SGD) performs slightly worse and parallel dither performs comparably but learns much more rapidly. The combination of parallel dither w/ dropout shows almost as much improvement over the regularised batch-SGD performance as the regularised batch-

SGD performance shows over the un-regularised batch-SGD (size: 32). Thus, it is clear that the non-batch SGD results with either parallel dither or parallel dither w/ dropout regularisation do substantially improve on the best batch-SGD results.

In summary, without regularisation non-batch SGD didn't work very well (substantially worse than batch-SGD), suggesting that batch averaging does indeed act as regularisation in its own right (mediated by the data). Both dither and dropout failed to improve things. However, parallel dither and parallel dither w/ dropout substantially improved on the batch-SGD results. These results suggest that with suitable regularisation, such as that described here, non-batch-SGD may be more useful than was previously thought. In particular, since the computation costs of the parallel regularisation described here may be absorbed by a parallel computing architecture, the promise of parallel regularisation is not just enhanced performance at test time but greatly reduced training time.

## IV. Discussion and Conclusion

In this paper, following the discrete signal processing interpretation of deep neural networks [5,6,2], we have demonstrated a robust new parallel regularisation method for training deep neural networks. In doing so, we have also challenged the common assumption that batch-averaged SGD is the best way to go. We have also offered some evidence that dither and dropout act on different components of the regularisation problem and that a combination of the two offers the best performance.

## Acknowledgment

AJRS did this work on the weekends and was supported by his wife and children.

## References

[1] Hinton GE, Srivastava N, Krizhevsky A, Sutskever I, Salakhutdinov R (2012) "Improving neural networks by preventing co-adaptation of feature detectors", The Computing Research Repository (CoRR), abs/1207.0580.

[2] Simpson AJR (2015) "Dither is Better than Dropout for Regularising Deep Neural Networks", arxiv.org abs/ 1508.04826.

[3] LeCun Y, Bottou L, Bengio Y, Haffner P (1998) "Gradient-based learning applied to document recognition", Proc. IEEE 86: 2278–2324.

[4] Hinton GE, Osindero S, Teh Y (2006). "A fast learning algorithm for deep belief nets", Neural Computation 18: 1527–1554.

[5] Simpson AJR (2015) "Abstract Learning via Demodulation in a Deep Neural Network", arxiv.org abs/1502.04042

[6] Simpson AJR (2015) Over-Sampling in a Deep Neural Network, arxiv.org abs/1502.03648.